\documentclass{article} 
\usepackage{subcaption}
\usepackage[final]{neurips_2023}
\usepackage{microtype}
\usepackage{hyperref}
\usepackage{url}
\usepackage{booktabs}
\usepackage[dvipsnames]{xcolor}
\usepackage{float}
\usepackage{makecell}
\usepackage{lineno}
\usepackage[most]{tcolorbox}

\definecolor{darkblue}{rgb}{0, 0, 0.5}
\hypersetup{colorlinks=true, citecolor=darkblue, linkcolor=darkblue, urlcolor=darkblue}

\usepackage{algorithm}
\usepackage{algorithmic}
\usepackage{amsmath}
\usepackage{graphicx}
\usepackage{fancyhdr}
\renewcommand{\headrulewidth}{0pt}
\title{Calibrated Uncertainty Quantification for Multi-Modal LLMs}
\title{Calibrating Uncertainty Quantification of Multi-Modal LLMs using Grounding}

\author{
 \textbf{Trilok Padhi* \textsuperscript{1}}
 \textbf{Ramneet Kaur* \textsuperscript{2}},
 \textbf{Adam D. Cobb\textsuperscript{2}},
  \textbf{Manoj Acharya\textsuperscript{2}},
 \textbf{Anirban Roy\textsuperscript{2}}, \\
  \textbf{Colin Samplawski\textsuperscript{2}},
 \textbf{Brian Matejek\textsuperscript{2}},
 \textbf{Alexander M. Berenbeim\textsuperscript{3}},
 \textbf{Nathaniel D. Bastian\textsuperscript{3}},\\
 \textbf{Susmit Jha\textsuperscript{2}}
\\
 \textsuperscript{1}Georgia State University, Atlanta, USA \\
 \textsuperscript{2}Computer Science Lab, SRI, Menlo Park, USA \\
 \textsuperscript{3}Army Cyber Institute, United States Military Academy, West Point, NY USA
\\
 \small{
   \textbf{Correspondence:} \href{mailto:email@domain}{tpadhi1@student.gsu.edu, ramneet.kaur@sri.com}
 }
}

\begin{document}

\maketitle

\thispagestyle{fancy}

\begin{abstract}

We introduce a novel approach for calibrating uncertainty quantification (UQ) tailored for multi-modal large language models (LLMs). Existing state-of-the-art UQ methods rely on consistency among multiple responses generated by the LLM on an input query under diverse settings. However, these approaches often report higher confidence in scenarios where the LLM is consistently incorrect. This leads to a poorly calibrated confidence with respect to accuracy. To address this, we leverage cross-modal consistency in addition to self-consistency to improve the calibration of the multi-modal models. Specifically, we ground the textual responses to the visual inputs. The confidence from the grounding model is used to calibrate the overall confidence. 
Given that using a grounding model adds its own uncertainty in the pipeline, we apply temperature scaling -- a widely accepted parametric calibration technique -- to calibrate the grounding model’s confidence in the accuracy of generated responses. We evaluate the proposed approach across multiple multi-modal tasks, such as medical question answering (Slake) and visual question answering (VQAv2), considering multi-modal models such as LLaVA-Med and LLaVA. The experiments demonstrate that the proposed framework achieves significantly improved calibration on both tasks. 
\end{abstract}

\section{Introduction}
\label{sec:intro}
Recent advancements in large language models (LLMs) have demonstrated their impressive performance across many
 domains, ranging from natural language processing~\citep{bert} and machine translation~\citep{machine_translation} to creative writing~\citep{create_writing} and code generation~\citep{jiang2024cursorcore}. Despite their capabilities, these models are not infallible and are known to produce incorrect or misleading information, often referred to as hallucinations~\citep{hallucination_survey}. 
Uncertainty Quantification (UQ) of LLMs has been proposed as a practical solution to assess trust in these models, particularly for their deployment in safety-critical areas such as healthcare~\citep{uq_survey}.
UQ techniques aim to provide a quantitative measure of trust that a user can place in an LLM's response to the input query.

State-of-the-art approaches for quantifying the uncertainty of LLMs are motivated by self-consistency theory~\citep{self_consistency_theory}. This involves prompting the model multiple times for the same input under diverse settings, such as with a high-temperature value, and checking for similarity in the generated responses for assessing model's uncertainty on the input~\citep{gen_with_confidence, kuhn, self_prob, kaur2024addressing}. The underlying idea is that if the model generates semantically similar responses for the same input under diverse settings, then it is certain (or confident) about the input. 

\begin{figure}[!t]
    \centering
    \includegraphics[width=1\linewidth]{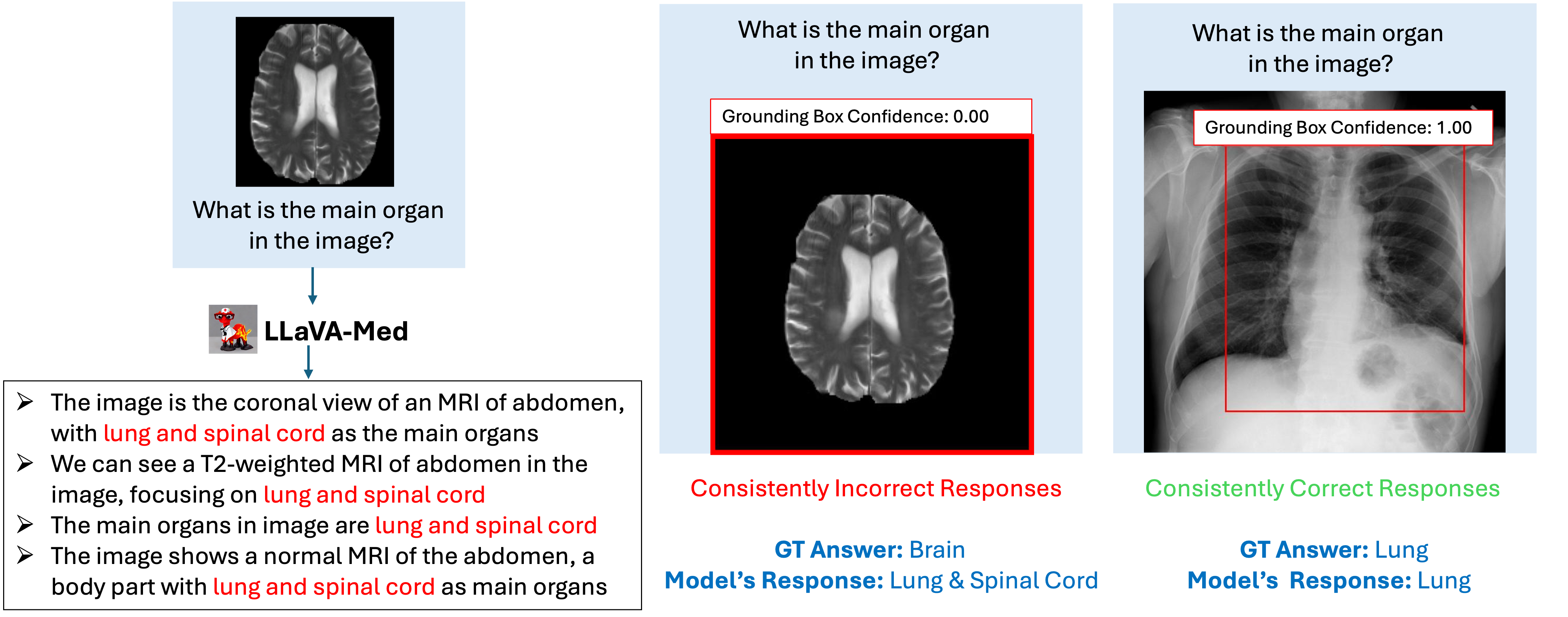}
    \caption{Consistently incorrect responses generated by LLaVA-Med-v1.5-Mistral-7B~\citep{li2023llava} on MRI image of the brain from the Slake Medical dataset (left). BiomedParse~\citep{biomedparse}, a grounding model for medical images, is not able to locate `lung \& spinal cord' on the MRI image of brain, and therefore labels the entire image as lung with zero confidence. It is, however, able to generate a bounding box for lung on the chest X-ray with $100\%$ confidence (right).}
    \label{fig:consistently_incorrect_responses}
\end{figure}

Consistency, however, does not imply accuracy. As shown in Fig.~\ref{fig:consistently_incorrect_responses} (left), we observe that models can generate consistently incorrect responses. This observation is made on $25$ inputs out of $75$ manually verified test cases on Slake, a medical dataset~\citep{liu2021slake}. The goal of UQ approaches is to provide a measure of confidence in the real-world performance of LLMs for their trustworthy deployment. The confidence in an LLM as reported by UQ approaches should, therefore, be aligned with the accuracy of the LLM. This alignment can be checked by plotting the expected accuracy of the model as a function of the reported confidence. These plots are known as \textit{reliability diagrams} and have been used to report confidence-accuracy calibration (or alignment) of deep learning models~\citep{guo2017calibration}.

\begin{figure*}[!t]
    \centering
    \includegraphics[width=1\linewidth]{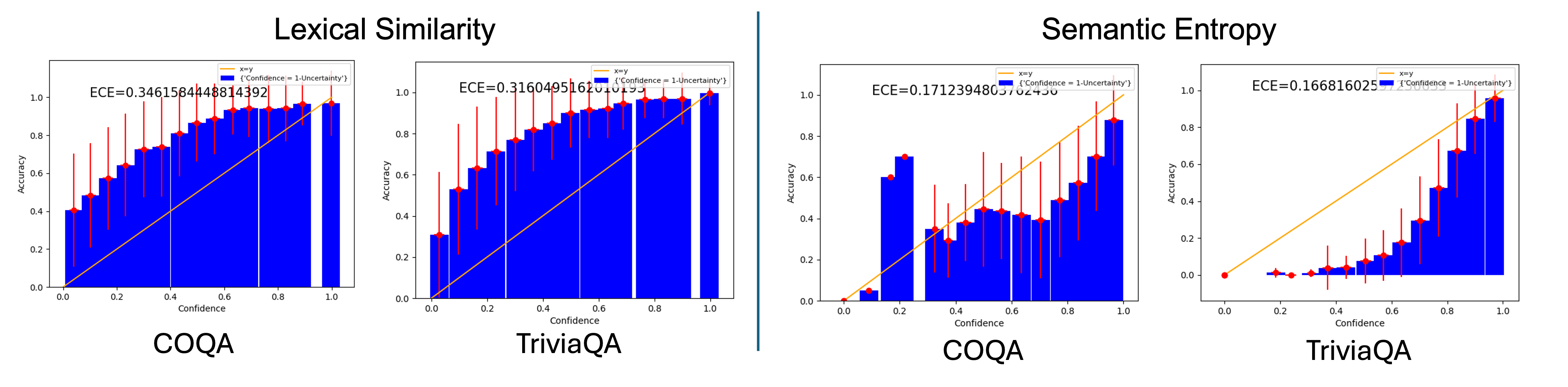}
    \caption{Reliability diagrams with the expected accuracy of Llama-2-13B on COQA and TriviaQA datasets plotted as a function of the model's confidence predicted by self-consistency-based UQ approaches: `Lexical Similarity'~\citep{lexical_sim} on the left and `Semantic Entropy'~\citep{kuhn} on the right.  A perfect calibration between the model's accuracy and the predicted confidence would have resulted in red points (average accuracy for each confidence bin) on the $x=y$ axis with a low variance (length of red lines).}
    \label{fig:cali_curves_llama_13B}
\end{figure*}

Fig.~\ref{fig:cali_curves_llama_13B} shows reliability diagrams for two self-consistency-based UQ approaches for predicting confidence of Llama-2-13B~\citep{llama-2} on COQA~\citep{coqa} and TriviaQA~\citep{triviaqa} datasets; a common test setting considered by these approaches~\citep{gen_with_confidence, kuhn, kaur2024addressing}. High expected calibration error (ECE) shows poor calibration of the model's accuracy with its confidence predicted by these approaches\footnote{GPT-4-Turbo~\citep{gpt} is used to report accuracy of responses by Llama-2-13B w.r.t the ground truth}. We observe similar trends of high ECE by self-consistency-based UQ approaches for LLMs when applied to the multi-modal input settings of text-image pairs. These observations are reported in the experimental section.


We propose an approach for estimating calibrated confidence for multi-modal LLMs. The goal is to improve the calibration of the confidence estimated by self-consistency-based UQ approaches by leveraging the consistency between the responses in multiple modalities. Specifically, in addition to checking the self-consistency between multiple textual responses, we ground the responses to the visual modality of the input query. For example, as shown in Fig.~\ref{fig:consistently_incorrect_responses} (right), we ground the answers with bounding boxes on the image. A correct answer has a higher chance of being grounded as the evidence is likely to be present in the image. Thus, the ability (or inability) to ground the generated response to the input image provides evidence about the correctness (or incorrectness) of the response.



\begin{figure*}[!t]
    \centering
    \includegraphics[width=1\linewidth]{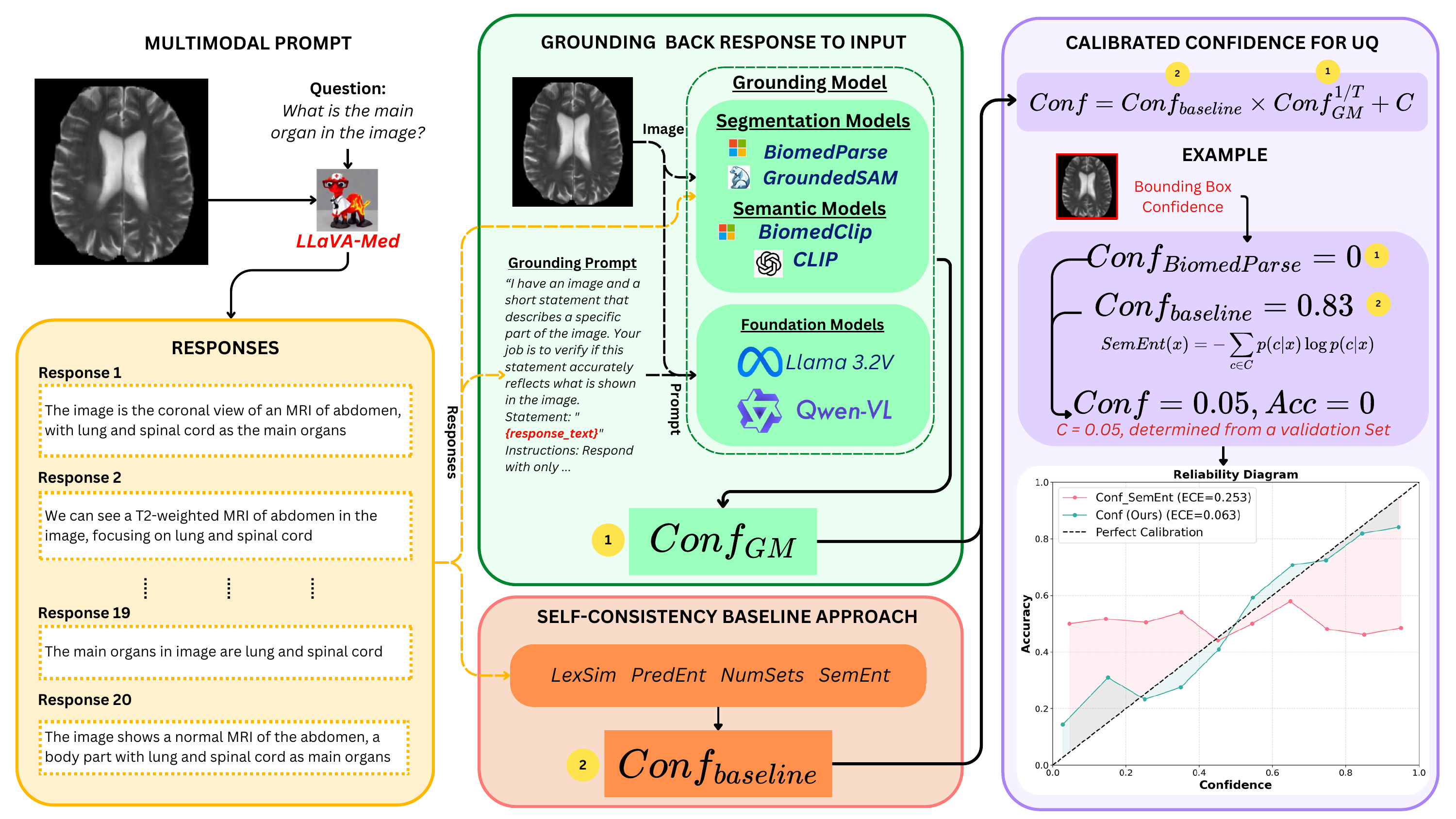}
    \caption{$Conf_{baseline}$: confidence from a self-consistency UQ baseline such as LexSim~\citep{lexical_sim}, PredEnt~\citep{pred_entropy}, NumSets~\citep{kuhn, gen_with_confidence}, SemEnt~\citep{kuhn}, etc. on a multi-modal LLM such as LLaVA-Med. $Conf_{GM}^{1/T}$: temperature-scaled calibrated confidence of grounding model on the accuracy of the generated responses. A grounding model can be as simple as the CLIP-based model, that provides its confidence in terms of a similarity score between embeddings of the generated response and the input image (e.g. BiomedCLIP), a detection model for response on the input image reporting its confidence on the detected bounding box (e.g. BiomedParse), or a foundation model that provides its verdict -- a confidence score in $[0,1]$ -- on the relevance of the generated response to the input image (e.g. LLaMA 3.2V, Qwen VL, etc.). $Conf$~\eqref{our_proposal} is the proposed calibrated confidence score for UQ of multi-modal LLMs resulting in substantially lower ECE than the baseline.}
    \label{fig:proposed_approach}
\end{figure*}

One consideration of relying on a grounding model to report the uncertainty of another model is that it introduces the grounding model's uncertainty into the pipeline. We apply temperature scaling to calibrate the grounding model's confidence in the accuracy of the multi-modal LLM. Temperature scaling is a simple yet effective post-processing parametric calibration technique that has been widely used to align the confidence with the true likelihood of correctness~\citep{temp_scaling_use2, temp_scaling_use1, guo2017calibration}. Fig.~\ref{fig:proposed_approach} shows the proposed approach for calibrated confidence prediction for multi-modal LLMs. Experimental results on two open-set question-answering datasets namely Slake (medical)~\citep{liu2021slake} and VQA (general objects Visual Question Answering)~\citep{vqa} with LLaVA-Med and LLaVA as the multi-modal LLMs respectively demonstrate promising results of the proposed approach with a variety of grounding models.

\section{Related Work}
\label{sec:rel_work}
There has been recent work on utilizing information from different modalities of multi-modal inputs to LLMs for their interpretability.~\citet{owl_vit_explaination} make use of an open-world localization model (OWL-ViT) for projecting bounding box on the identified objects by the LLM. They do so by a joint training in the embedding space of the LLM and OWL-ViT. This approach is applicable to open-source models for joint training in the embedding space of vision modality.~\citet{pelican} run object detection model on an input image and make use of the detected objects to check the LLM's response for hallucinations. This is done by generating a claim from the query-response pair and then verifying the decomposed sub-claims against the detected objects.

There has been a recent focus on measuring uncertainty in multi-modal LLMs via self-consistency theory.~\citet{pert_modalities_uq} apply perturbations in both text and image input modalities, and use entropy in the distribution of the generated responses for reporting uncertainty of the LLM on the input query. 
This can be used as the self-consistency UQ baseline in the proposed framework for calibrating the reported confidence (by the baseline) in multi-modal LLMs. 
~\citet{li2024graph} propose to train a Graph Neural Network (GNN) on clusters of semantically equivalent responses to predict the probability of each response being correct, with the optimization goal of calibrating the predicted probabilities. This is, however, a supervised approach that requires a labeled training set for the GNN. Our approach does not require training any new model but involves calibrating the confidence of grounding models (GM) via temperature scaling on a small validation set to take into account the uncertainty of GM used in the pipeline. The use of a validation set is a common setting in uncertainty quantification literature such as conformal prediction~\citep{cp}, where the validation set\footnote{a validation set is known as the calibration set in conformal prediction framework.} is used to determine a threshold on the membership score (known as non-conformity score) of a generated response in the output set. The output is a set instead of a single prediction from the LLM to take into account uncertainty in the LLM with coverage guarantees on the set~\citep{wang2025copu, UQ_benchmark, kaur2024addressing, conformal_lang_modeling}.


Prompting the LLMs to generate a confidence score along with its prediction has also been used to quantify the model's uncertainty~\citep{self_prob, tian2023just, xiongcan}. High calibration error in the self-confidence by vision large language models (VLLMs) has been observed and reported in the literature~\citep{self_conf_cali_1, self_conf_cali_2}. The authors report over-confidence by VLLMs but do not propose any calibration technique for those. To the best of our knowledge, this is the first work on calibrating confidence from existing UQ approaches when applied to multi-modal LLMs.

\section{Background}
\label{sec:background}

\subsection{Uncertainty Quantification of LLMs} 
Uncertainty quantification (UQ) techniques aim to measure the uncertainty in the predictions of LLMs to assess their reliability. A prominent strategy for UQ is built on self-consistency theory~\citep{self_consistency_theory}, which focuses on generating multiple responses from the model for the same input and evaluating the consistency among these outputs. This strategy quantifies uncertainty by identifying discrepancies in the generated responses. Different approaches for UQ of LLMs differ in how they measure these discrepancies and the metrics they use to quantify the resulting uncertainty.

\textit{Predictive entropy} over the probability distribution of responses is a popular uncertainty metric, and serves as a baseline in UQ for LLMs~\citep{self_prob, kuhn, kaur2024addressing, gen_with_confidence}. In the context of natural language processing, for an input query $x$ the probability of a response $r$ is calculated as the product of the conditional log probabilities for each token in the response: $p(\mathbf{r}|x) = \prod_i p(r_i|r_{<i},x).$ \textit{Lexical similarity} (similar to~\citep{lexical_sim}) is another proposed approach that assigns a similarity score between each pair of responses $(r_i, r_j)$ via RougeL~\citep{rouge}\footnote{RougeL measures the similarity between two sentences based on the longest common subsequence between the two.}, and uses the average of this score over each pair as the confidence (1-UQ) metric in the LLM: $\frac{1}{P}\sum_{i=1}^{|R|}\sum_{j=1}^{|R|}\text{RougeL}(r_i, r_j)$. Here $R$ is the set of responses, and $P= |R|\times(|R|-1)/2$.~\citet{kuhn} group the generated responses into semantically equivalent response clusters $C$, compute the probability of each cluster as the average probability of each response in the cluster, and use \textit{semantic entropy} ($SE$) over these clusters as the uncertainty metric: $SE(x) = -\sum_{c \in C} p(c|x)\log p(c|x),$ Semantic equivalence between two responses for clustering them together is checked via bi-directional entailment between the two via Deberta (Natural Language Inference) model~\citep{he2020deberta}.~\citet{kaur2024addressing} further enhances this approach via a new dynamic semantic clustering algorithm for deciding on the membership of a response. Another UQ metric used as a baseline in~\citep{kuhn, kaur2024addressing} is \textit{NumSets}, that is the number of semantic clusters formed from the responses. 

\subsection{Confidence Calibration}
Confidence in a model's predictions is known to be calibrated if it represents the true probability of those predictions to be correct~\citep{guo2017calibration}. This means that for a set of predictions where the model's confidence estimate is $c$, the proportion of correct predictions by the model should be $c$.


\textit{Temperature Scaling}~\citep{guo2017calibration} is a simple yet effective post-processing parametric method for calibrating the softmax confidence of classification models. It learns a single parameter ($T$) to scale the model's logits by a factor of $1/T$ with greater values of $T$ softening the softmax distribution indicating high uncertainty, and lower values of $T$ sharpening the distribution indicating low uncertainty.


\textit{Reliability diagrams}~\citep{reliability_diagram}, such as the ones shown in Fig.~\ref{fig:cali_curves_llama_13B}, provide us with a way to visualize confidence calibration empirically. These diagrams plot the expected accuracy of the model as a function of confidence. The range of confidence in $[0, 1]$ is divided into equal-sized smaller bins, and accuracy for each bin is calculated as an average accuracy on the samples falling in that bin. A perfect calibration would result in average accuracy for each bin equal to the average confidence of the bin. Any deviation from this identity function indicates a miscalibration. \textit{Expected Calibration Error} (ECE)~\citep{ECE} provides us with a measure for this miscalibration. ECE is calculated as the weighted average over each bin's difference in accuracy and confidence~\citep{guo2017calibration}:
\begin{equation}
    \sum_{m=1}^{M}\frac{N_m}{n} |Acc_m - Conf_m|,
\end{equation}
where $M$ is the number of confidence bins, $N_m$ is the number of samples in bin $m$, $n$ is the total number of samples, and $Acc_m$, $Conf_m$ are the average accuracy and confidence of the $m^{th}$ bin.

\subsection{Grounding}
Grounding refers to the process of linking symbolic representations of knowledge (e.g. language) to the real world's sensory data (e.g., images, sounds). 
For example, visual grounding techniques~\citep{yu2018mattnet, acharya2019vqd} have been proposed for mapping textual entities to bounding boxes on images. Grounding models (GM) such as object detection or localization models have been used to report interpretability of LLMs~\citep{owl_vit_explaination, pelican}. We propose the use of grounding models for calibrating the UQ of multi-modal LLMs. Different types of GM can be utilized for linking back the generated response by an LLM to the input space. We consider three of those: \textit{segmentation-based GM} that generates a segmentation mask relevant to the generated response on the input image, \textit{semantic-based GM} that assigns a similarity score to the generated response and the input image, and \textit{foundation model} that can be instructed to verify the accuracy of the response as the description of the input image.

\section{Calibrated Uncertainty Quantification}
\label{sec:tech}
The core idea behind the proposed approach is to leverage cross-modal consistency for evaluating the accuracy of the generated responses. This inference time accuracy evaluation can be utilized for calibrating the UQ of multi-modal LLMs on the input query. The ability to ground back a response into the provided context generates evidence about the response's accuracy. Fig.~\ref{fig:proposed_approach} shows examples of grounding models (GM) that quantitatively provide this evidence by estimating a confidence score in the accuracy of LLM's response. We use this GM-generated confidence (a score in $[0, 1]$) to calibrate the confidence reported by the self-consistency-based UQ approaches. Specifically, confidence in a multi-modal LLM by a self-consistency baseline, $Conf_{baseline} \in [0,1]$, is calibrated by multiplying it with the GM's confidence in the accuracy of the LLM on the input query:
\begin{equation}
    Conf = Conf_{baseline} \times Conf_{GM}^{(1/T)} + C,
    \label{our_proposal}
\end{equation}
where $Conf_{GM}$ is the average confidence of a GM over all the responses by the LLM generated in diverse settings for self-consistency checking. $T (> 0)$ is the hyper-parameter for temperature scaling or calibration of $Conf_{GM}$.
Higher values of $T$ sharpen (or increase) the confidence of the grounding model. Reducing the value of $T$ reduces the confidence of the grounding model, and therefore the overall confidence. This is necessary to make sure the grounding model is calibrated correctly to the task. A constant $C (> 0)$ is also added to offset the reduction caused due to the product of confidences in the range of
 $[0, 1]$. Both hyperparameters $T$ and $C$ are determined from a validation set.



\section{Experiments}
\label{sec:exp}

\subsection{Datasets} We conduct experiments on two open-ended visual question answering (QA) datasets from different domains: \textbf{S}emantically-\textbf{la}beled
\textbf{k}nowledge-\textbf{e}nhance (Slake)~\citep{liu2021slake} dataset from the medical domain, and \textbf{V}isual \textbf{Q}uestion \textbf{A}nswering (VQA v2.0)~\citep{vqa} dataset from the general domain for testing commonsense visual knowledge. 

\textbf{Slake} is a bilingual (English and Chinese) dataset with $14K$ QA pairs on $642$ images with CT, MRI, and X-Ray as the different image modalities. The question type can be vision-only or knowledge-based on $12$ diseases and $39$ organs of the whole body with ground-truth labels from physicians. Fig.~\ref{fig:consistently_incorrect_responses} shows examples of images from the dataset. We filter the English QA pairs from the the test set of the bilingual Slake, and use $80\%$ as the test and $20\%$ as the validation set for our experiments.

\textbf{VQA} dataset contains general QA pairs that evaluates the visuo-linguistic understanding of models. We use VQAv2 version of the dataset since it discourages the model to solely rely on the language priors and encourages joint understanding of the image and query. Some examples of image-question pairs from this dataset are shown in Appendix. For our experiments, we use again use $80/20$ percentage split for test/validation splits on the test set of VQAv2. 

\subsection{Multi-Modal LLMs}
We consider LLaVA-v1.5-7B~\citep{llava_1.5} and LLaVA-Med-v1.5-Mistral-7B~\citep{llava_med} as the multi-modal LLMs for quantifying their uncertainty on VQA and Slake, respectively. \textbf{L}arge \textbf{L}anguage \textbf{a}nd \textbf{V}ision \textbf{A}ssistant (LLaVA) is a vision and language model that connects a vision encoder (CLIP~\citep{clip}) with a language model (LLaMA 2~\citep{llama-2}) to handle image-text queries.
LLaVA-Med-v1.5-Mistral-7B is the fine-tuned variant of LLaVA on the medical domain with Mistral-7B~\citep{mistral} as the language model. The fine-tuning is done by curriculum learning on biomedical image-caption pairs from the PubMed Central~\citep{pubmed_central} dataset.

\subsection{Grounding Models}
We consider different categories of the grounding models (GM). For VQA, we consider the following GM: 
\begin{enumerate}
    \item Segmentation-based GM: GroundedSAM~\citep{grounded_sam} trained to perform segmentation on the bounding box detected by Grounding DINO~\citep{grounding_dino} on an image for the input text.
    \item Semantic-based GM: CLIP~\citep{clip} is the \textbf{C}ontrastive \textbf{L}anguage–\textbf{I}mage \textbf{P}re-training model that assigns a similarity score between the image-text pair in their embedding space.
    \item Foundation GM: LLaMA-3.2-11B-Vision-Instruct~\citep{llama3.2v} (LLaMA3.2V), and Qwen2-VL-7B-Instruct~\citep{wang2024qwen2} (QwenVL). 
\end{enumerate}
For Slake, we report results with semantic and foundation GM:
\begin{enumerate}
    \item Semantic-based GM: BiomedClip~\citep{zhang2023biomedclip}, an advanced version of CLIP model fine-tuned on the medical domain.
    \item Foundation GM: Biomed-Qwen2-VL-2B-Instruct~\citep{med_qwen} (Biomed-QwenVL), the fine-tuned version of QwenVL on medical domain, and LLaMA-3.2-11B-Vision-Instruct (LLaMA3.2V)\footnote{To the best of our knowledge, there is no model LLaMA3.2V family of foundation models specific to the medical domain}. 
\end{enumerate}

We add details about the prompts for foundation GM in the Appendix.



\begin{table}[!t]
    \centering
    \renewcommand{\arraystretch}{1.2}
    \setlength{\tabcolsep}{5pt}
    \resizebox{\columnwidth}{!}{
    \begin{tabular}{l cc cccc}
        \toprule
        \raisebox{-2ex}{\textbf{Baseline}}  
            & \multicolumn{2}{c}{\textbf{Without Grounding} ($\downarrow$)} 
            & \multicolumn{4}{c}{\textbf{With Grounding} ($\downarrow$)} \\
        \cmidrule(lr){2-3} \cmidrule(lr){4-7}
            & $Conf_{\text{baseline}}$ 
            & $Conf_{\text{baseline}}^{(1/T)}$ 
            & GroundedSAM 
            & CLIP 
            & LLaMA3.2V 
            & QwenVL \\
        \midrule
        \textbf{LexSim}    
            & 0.169 
            & 0.081 
            & 0.052 \(\,(\textcolor{ForestGreen}{-36.0\%})\) 
            & \textbf{0.045} \(\,(\textcolor{ForestGreen}{-44.6\%})\) 
            & \textbf{0.045} \(\,(\textcolor{ForestGreen}{-44.6\%})\) 
            & 0.061 \(\,(\textcolor{ForestGreen}{-25.0\%})\) \\
        \textbf{NumSets}   
            & 0.410 
            & 0.246 
            & 0.143 \(\,(\textcolor{ForestGreen}{-41.9\%})\) 
            & 0.143 \(\,(\textcolor{ForestGreen}{-41.8\%})\) 
            & 0.127 \(\,(\textcolor{ForestGreen}{-48.4\%})\) 
            & \textbf{0.125} \(\,(\textcolor{ForestGreen}{-49.2\%})\) \\
        \textbf{PredEnt}   
            & 0.445 
            & 0.190 
            & 0.115 \(\,(\textcolor{ForestGreen}{-39.6\%})\) 
            & 0.117 \(\,(\textcolor{ForestGreen}{-38.3\%})\) 
            & \textbf{0.096} \(\,(\textcolor{ForestGreen}{-49.4\%})\) 
            & 0.119 \(\,(\textcolor{ForestGreen}{-37.5\%})\) \\
        \textbf{SemEnt}    
            & 0.108 
            & 0.108 
            & 0.036 \(\,(\textcolor{ForestGreen}{-66.9\%})\) 
            & 0.038 \(\,(\textcolor{ForestGreen}{-64.6\%})\) 
            & \textbf{0.029} \(\,(\textcolor{ForestGreen}{-73.1\%})\) 
            & 0.073 \(\,(\textcolor{ForestGreen}{-32.5\%})\) \\
        \bottomrule
    \end{tabular}
    }
    \caption{Comparison of ECE over accuracy of LLaVA for VQA with the confidence reported by baseline ($Conf_{baseline}$), calibrated baseline ($Conf_{baseline}^{(1/T)}$), and the proposed calibration with grounding~\eqref{our_proposal}. Percentage improvement in ECE via grounding from the calibrated baseline is also reported (in green) for each grounding model.}
    \label{tab:vqa_random_combined}
\end{table}
\begin{table}[!t]
    \centering
    \renewcommand{\arraystretch}{1.2}
    \setlength{\tabcolsep}{5pt}
    \resizebox{\columnwidth}{!}{
    \begin{tabular}{l cc ccc}
        \toprule
        \raisebox{-2ex}{\textbf{Baseline}} 
            & \multicolumn{2}{c}{\textbf{Without Grounding} ($\downarrow$)} 
            & \multicolumn{3}{c}{\textbf{With Grounding} ($\downarrow$)} \\
        \cmidrule(lr){2-3} \cmidrule(lr){4-6}
            & $Conf_{\text{baseline}}$ 
            & $Conf_{\text{baseline}}^{(1/T)}$ 
            & BiomedClip 
            & LLaMA3.2V 
            & Biomed-QwenVL \\
\midrule
LexSim  & 0.031 & 0.031 & \textbf{0.004} \(\,(\textcolor{ForestGreen}{-87.1\%})\) & 0.013 \(\,(\textcolor{ForestGreen}{-58.1\%})\) & 0.038 \(\,(\textcolor{red}{+22.6\%})\) \\
NumSets & 0.426 & 0.201 & \textbf{0.007} \(\,(\textcolor{ForestGreen}{-96.5\%})\) & 0.021 \(\,(\textcolor{ForestGreen}{-89.6\%})\) & 0.132 \(\,(\textcolor{ForestGreen}{-34.3\%})\) \\
PredEnt & 0.390 & 0.215 & \textbf{0.007} \(\,(\textcolor{ForestGreen}{-96.7\%})\) & 0.014 \(\,(\textcolor{ForestGreen}{-93.5\%})\) & 0.217 \(\,(\textcolor{red}{+00.9\%})\) \\
SemEnt  & 0.376 & 0.222 & \textbf{0.008} \(\,(\textcolor{ForestGreen}{-96.4\%})\) & 0.010 \(\,(\textcolor{ForestGreen}{-95.5\%})\) & 0.247 \(\,(\textcolor{red}{+11.3\%})\) \\
\bottomrule
    \end{tabular}
    }
    \caption{Comparison of ECE over accuracy of LLaVA for VQA with the confidence reported by baseline ($Conf_{baseline}$), calibrated baseline ($Conf_{baseline}^{(1/T)}$), and the proposed calibration with grounding~\eqref{our_proposal}. Percentage improvement/regression in ECE via grounding from the calibrated baseline is also reported (in green/red) for each grounding model.}
    \label{tab:slake_random_combined}
\end{table}

\subsection{Baselines}
We consider all four self-consistency-based UQ baselines described in the background section: Predictive Entropy (PredEnt), Lexical Similarity (LexSim), Semantic Entropy (SemEnt), and NumSets. The multi-modal LLM is prompted $20$ times for generating multiple responses required by the baselines under diverse input settings for the LLM with temperature $=0.5$ for randomness, and top\_p $=1$ for nucleus sampling. For a fair comparison, we also consider calibrating all baselines directly with temperature scaling denoted as $Conf_{baseline}^{(1/T)}$ where $T$ is learned using the validation set.

\subsection{Results}
We report the average \textit{Expected Calibration Error} (ECE) over accuracy with the confidence by (a) self-consistency baselines, (b) calibrated version of these baselines with temperature scaling, and (c) proposed calibration~\eqref{our_proposal} for these baselines via grounding for different grounding models. Average ECE is calculated from $5$ runs of random splits for the test/validation sets. Tables~\ref{tab:vqa_random_combined}, and~\ref{tab:slake_random_combined} show these results for VQA and Slake respectively. We observe a very low variance in all the cases, and it is included in the Appendix. Values of hyperparameters ($T$, and $C$) are also reported in the Appendix.

\begin{figure*}[!t]
    \centering
    \includegraphics[width=1\linewidth]{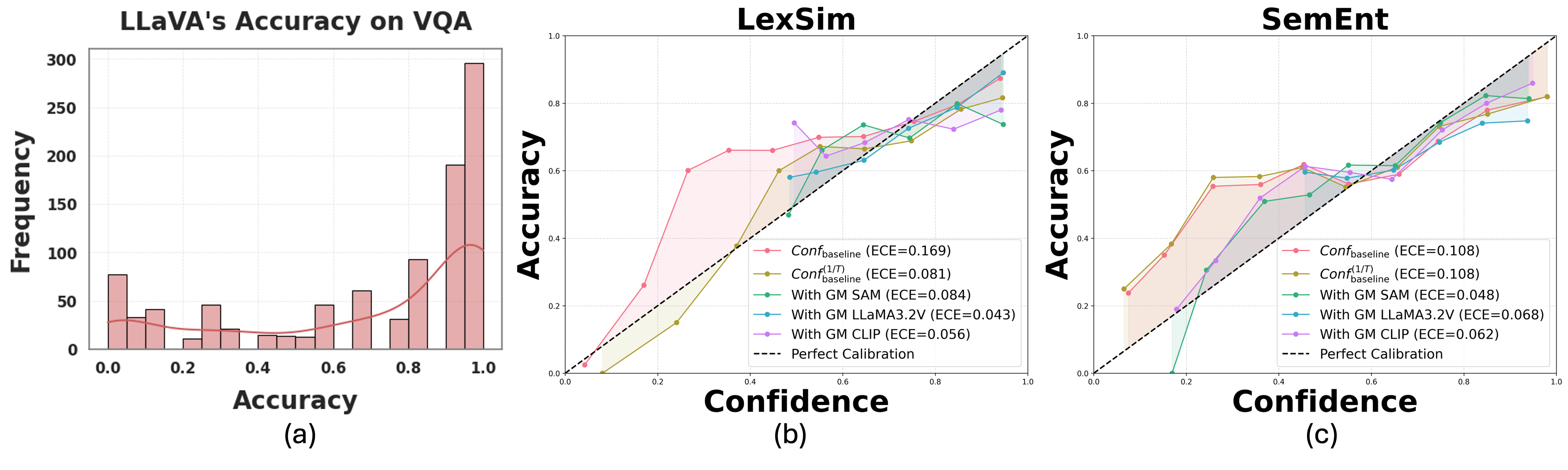}
    \caption{(a) Histogram on the frequency of LLaVA's accuracy on VQA. Reliability Diagrams for UQ of LLaVA on VQA by (b) LexSim, and (c) SemEnt baseline. Each diagram shows plots and the respective ECE for the confidence reported by the baseline $Conf_{baseline}$, its calibrated version $Conf_{baseline}^{(1/T)}$, and the proposed approach~\eqref{our_proposal} for calibration with different grounding models.}
    \label{fig:rel_dia_vqa}
\end{figure*}

\begin{figure*}[!t]
    \centering
    \includegraphics[width=1\linewidth]{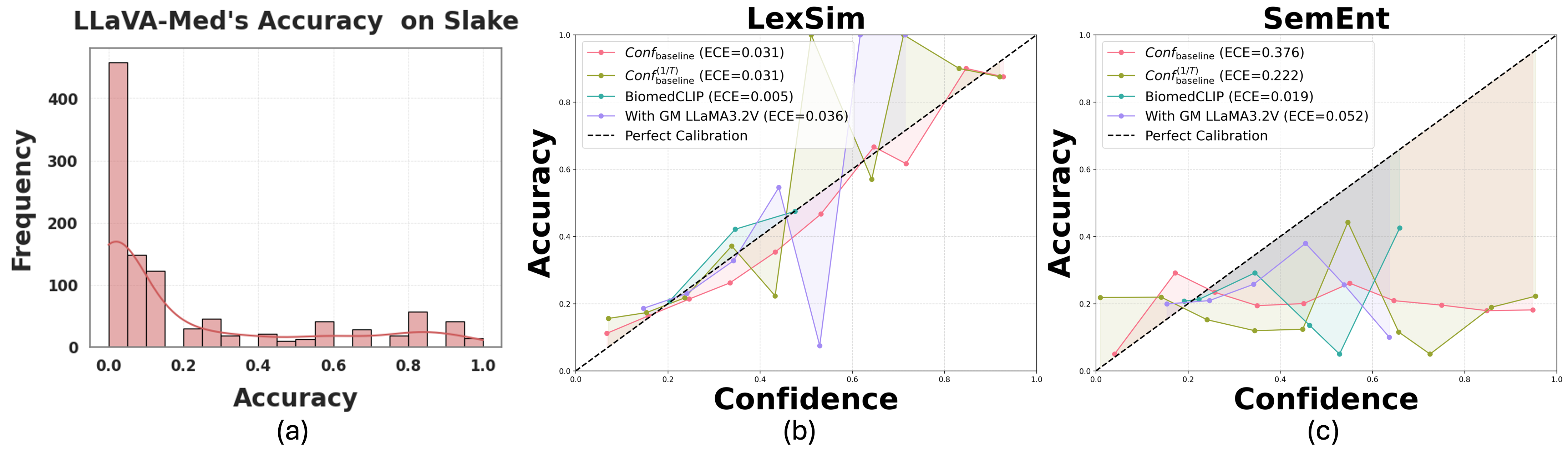}
    \caption{(a) Histogram on the frequency of LLaVA-Med's accuracy on Slake. Reliability Diagrams for UQ of LLaVA-Med on Slake by (b) LexSim, and (c) SemEnt. Each diagram shows plots and the respective ECE for the confidence reported by the baseline $Conf_{baseline}$, its calibrated version $Conf_{baseline}^{(1/T)}$, and the proposed approach~\eqref{our_proposal} for calibration with different grounding models.}
    \label{fig:rel_dia_slake}
\end{figure*}

We also plot \textit{reliability diagrams} for these baselines along with their temperature scaled version, and the proposed grounding approach for calibration of these baselines with different grounding models. These plots along with their respective ECE for LexSim, and SemEnt are reported in Fig.~\ref{fig:rel_dia_vqa} for VQA, and~\ref{fig:rel_dia_slake} for Slake: LexSim and Sement are the top two UQ baselines in terms of ECE for both the datasets. Plots for the other two baselines (PredEnt and NumSets) on both datasets are included in the Appendix. 

\paragraph{Observations on ECE:} We make the following observations from Tables~\ref{tab:vqa_random_combined} and ~\ref{tab:slake_random_combined}. First, 
ECE by the calibrated confidence via temperature scaling of all baselines is much lower than their original versions reported in the literature. This illustrates the efficacy of temperature scaling in calibrating the existing UQ techniques.
Second, compared to confidence by the self-consistency approaches ($Conf_{baseline}$), the proposed grounding-based approach achieves much lower ECE in all but one test case for all grounding models on both VQA and Slake datasets illustrates that the proposed approach is agnostic to the choice of GM.  Finally, significant percentage improvements in ECE in most (all but three as shown in red) test cases - at least $32.5\%$ for VQA and $34.3\%$ for for Slake - from the calibrated version of the baselines (as shown in green) illustrates that calibrating confidence of self-consistency scores on LLM's responses with external grounding is much more effective than calibrating these confidence scores via temperature scaling.

Another important observation is that the amount of ECE improvement depends on the choice of the GM. In the case of VQA, LLaMA3.2V performs the best. 
Our hypothesis on this is as follows. LLaMA3.2V's diverse pretraining corpus has enhanced the model's capacity for commonsense reasoning and nuanced interpretation of complex image-text relationships, making it apt for grounding the responses of the VQA dataset that requires commonsense knowledge and understanding of visual domain.
In the case of Slake, BiomedClip which is fine-tuned on medical domain performs better than the general-purpose LLaMA3.2V model. Although, Biomed-QwenVL is also fine-tuned for medical purposes but the use of synthetic data in the post-training might indicate its poor performance in the three baselines. 

\paragraph{Observations on Reliability Diagrams:} 
For VQA, LLaVA is able to answer most of the questions accurately - this is evident from the histogram on the frequency of LLaVA's accuracy on VQA~\ref{fig:rel_dia_vqa} (a). This justifies the concentration of the reliability plots in the higher accuracy-confidence range ($>0.5$) with all GM in case of LexSim and with LLaMA3.2V (best GM) on SemEnt. For other GM (SAM \& CLIP) on SemEnt, the plot is more calibrated -- closer to $x=y$ axis in all regions -- in comparison to both the baseline and the temperature-scaled baseline. 

For Slake, LLaVA-Med is not able to answer most of the questions accurately - this is evident from the histogram on the frequency of LLaVA-Med's accuracy on Slake~\ref{fig:rel_dia_slake} (a). This justifies the concentration of the reliability plots in the lower accuracy-confidence range ($< 0.5$) BiomedClip (best GM) on LexSim. The peaky behavior of the calibrated baseline and LlaMA3.2V in LexSim is due to uncalibrated confidence predictions on a very small number ($2$ to $3$) inputs from higher ($> 0.5$) confidence bins. Similarly in case of SemEnt, BiomedLCip and LLaMA3.2V yields more calibrated curve in the lower accuracy-confidence range ($< 0.5$) but again we observe peaky behavior here, again due to uncalibrated confidence predictions on $2$ to $3$ inputs from higher confidence bins.

We report similar results with the other two baselines (PredEnt and NumSets) for both VQA and Slake in the Appendix.

\section{Conclusion}
\label{sec:con}
This work sheds light on the limitations in the current state-of-the-art uncertainty quantification approaches for LLMs based on self-consistency theory, underlining an important distinction: consistency does not imply accuracy. Relying solely on consistency can, therefore, be misleading with incorrect confidence estimation in the LLM. We propose a nuanced approach on calibrated UQ for multi-modal LLMs that leverages cross-modal response consistency in addition to self-consistency by the existing approaches. Experimental results in different domains advocates the efficacy of the proposed approach. In future, in addition to image-text input modalities, we plan to extend our approach to other modalities such as image-audio, video-text, and video-audio.

\section*{Acknowledgments}
This material is based on work supported by the United States Air Force and Defense Advanced Research Projects Agency (DARPA) under Contract No.FA8750-23-C-0519, the Defense Advanced Research Projects Agency (DARPA) under Agreement No. HR0011-24-9-0424, and the Advanced Research Projects Agency for Health (ARPA-H) under Contract Number SP4701-23-C-0073. Any opinions, findings and conclusions or recommendations expressed in this material are those of the authors and do not necessarily reflect the United States Air Force, Department of Defense, Defense Advanced Research Projects Agency (DARPA), Advanced Research Projects Agency for Health (ARPA-H) or the United States Government. Research was sponsored by the Army Research Laboratory and was accomplished under Cooperative Agreement Number W911NF-17-2-0196.  The views and conclusions contained in this document are those of the authors and should not be interpreted as representing the official policies, either expressed or implied, of the Army Research Laboratory or the U.S. Government. The U.S. Government is authorized to reproduce and distribute reprints for Government purposes notwithstanding any copyright notation herein.




\newpage
\bibliography{colm2025_conference}
\bibliographystyle{colm2025_conference}

\appendix
\section{Appendix}

\subsection{Prompt for the Foundation Grounding Models}

\begin{tcolorbox}[colback=gray!10, colframe=black, width=\textwidth, sharp corners, title=\textbf{Prompt}]
\ttfamily
"I have an image and a short statement that describes a specific part of the image. Your job is to verify if this statement accurately reflects what is shown in the image.

Image: \textless Attached above\textgreater

Statement: ``\{response\_text\}''

Instructions: Respond with only one word — either ``Yes'' if the statement is correct, ``No'' if the statement is incorrect, or ``Not sure'' if you are uncertain. Do not provide any additional explanations."
\end{tcolorbox}

We map ``Yes'' to the confidence score of $1$, ``No'' as well as ``Not sure'' to the confidence score of $0$.

\subsection{Examples of Question-Answer Pairs from VQAv2 Dataset~\citep{vqa}}
\begin{figure}[H]
    \centering
    \begin{subfigure}[t]{0.45\textwidth}
        \includegraphics[width=\linewidth, height=4cm]{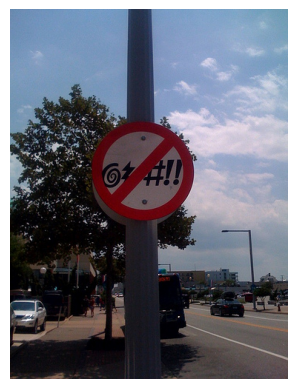}
        \caption*{\textbf{Q1:} What does the sign say?\\\textbf{A:} 'no cursing'}
    \end{subfigure}
    \hfill
    \begin{subfigure}[t]{0.45\textwidth}
        \includegraphics[width=\linewidth, height=4cm, keepaspectratio]{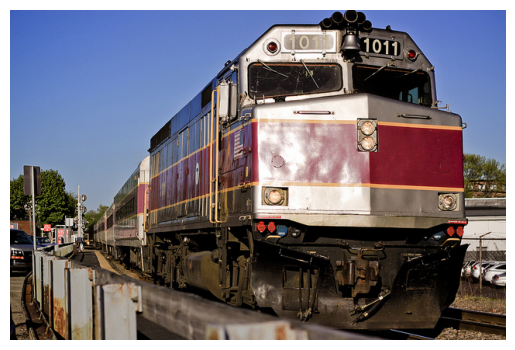}
        \caption*{\textbf{Q2:} What color is the engine?\\\textbf{A:} Red}
    \end{subfigure}

    \vspace{1em}

    \begin{subfigure}[t]{0.45\textwidth}
        \includegraphics[width=\linewidth, height=4cm]{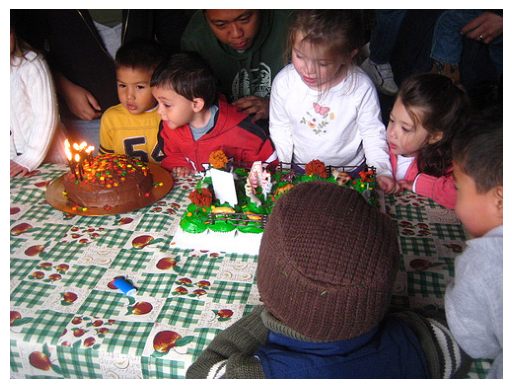}
        \caption*{\textbf{Q3:} Are most of the people wearing hats?\\\textbf{A:} No}
    \end{subfigure}
    \hfill
    \begin{subfigure}[t]{0.45\textwidth}
        \includegraphics[width=\linewidth, height=4cm, keepaspectratio]{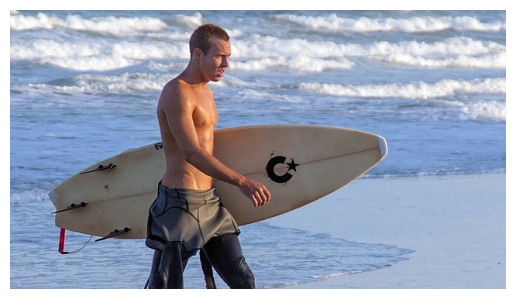}
        \caption*{\textbf{Q4:} Who does the man on the right resemble?\\\textbf{A:} 'surfer'}
    \end{subfigure}
    \caption{Example questions and answers from the VQA dataset.}
\label{app:vqa_image_samples}
\end{figure}

\subsection{Reliability Diagrams}
\vspace{-4em}  

\begin{figure}[H]
    \centering
    \begin{subfigure}[b]{0.45\textwidth}
        \includegraphics[width=\textwidth]{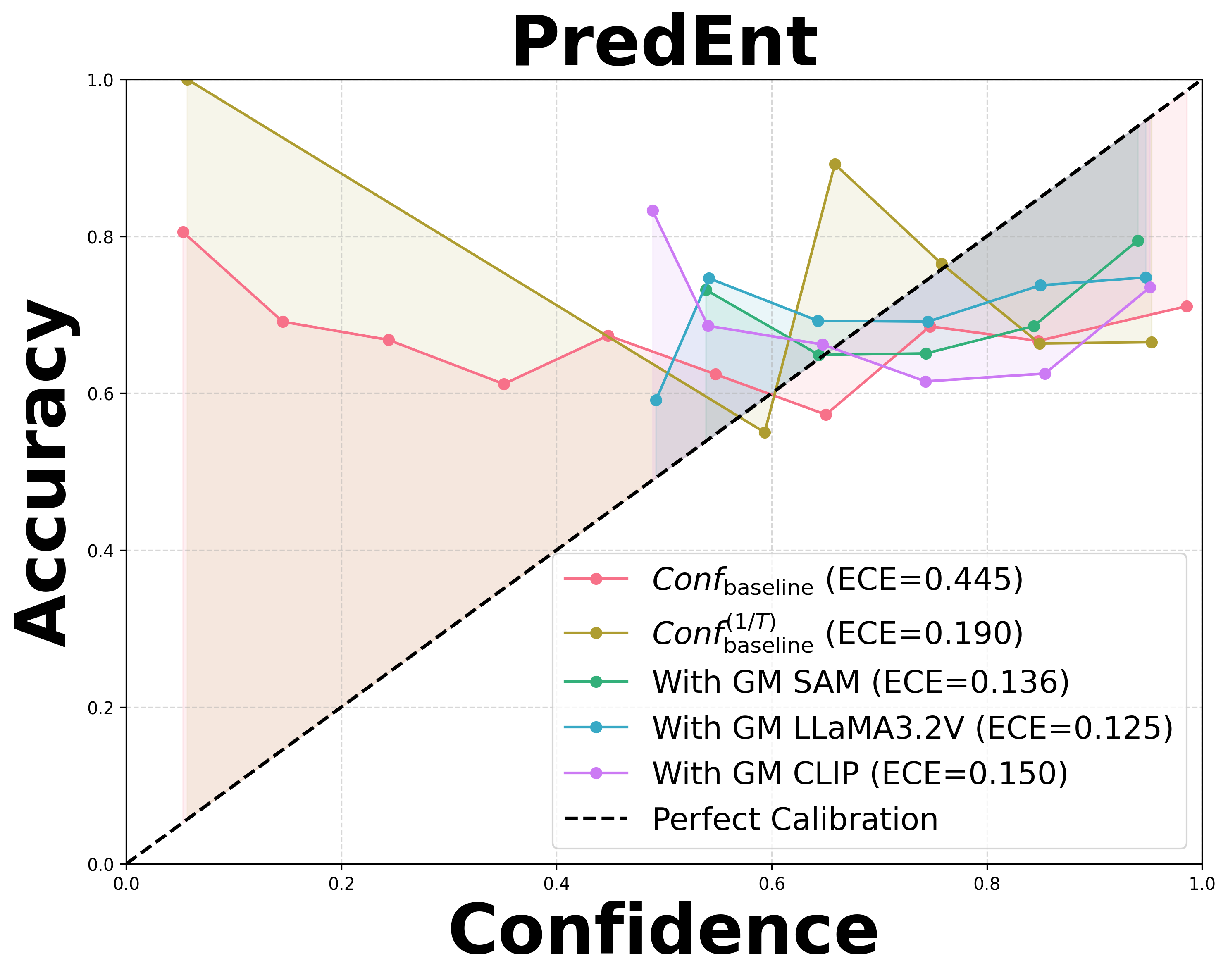}
        \label{fig:pred_entropy}
    \end{subfigure}
    \begin{subfigure}[b]{0.45\textwidth}
        \includegraphics[width=\textwidth]{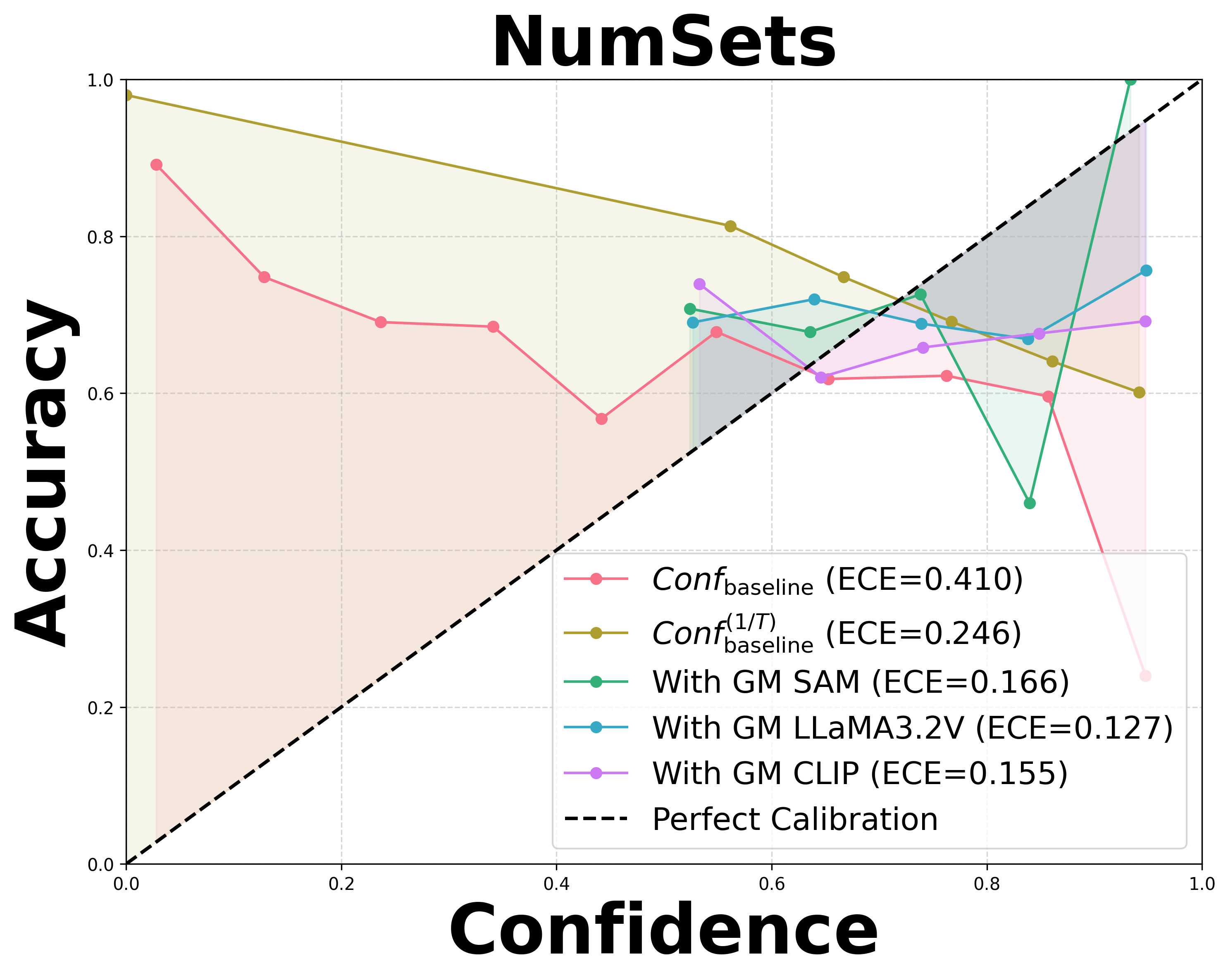}
        \label{fig:num_clusters}
    \end{subfigure}
    \caption{Reliability Diagrams for UQ by the four self-consistency baselines (PredEnt, SemEnt, LexSim, NumSets) of LLaVA on the VQA dataset. Each diagram shows plots and ECE for the confidence reported by the baseline $Conf_{baseline}$, its calibrated version $Conf_{baseline}^{(1/T)}$, and the proposed approach~\eqref{our_proposal} for calibration with different grounding models.}
    \label{fig:reliability_diagrams_vqa}
\end{figure}

\begin{figure}[H]
    \centering
    \begin{subfigure}[b]{0.45\textwidth}
        \includegraphics[width=\textwidth]{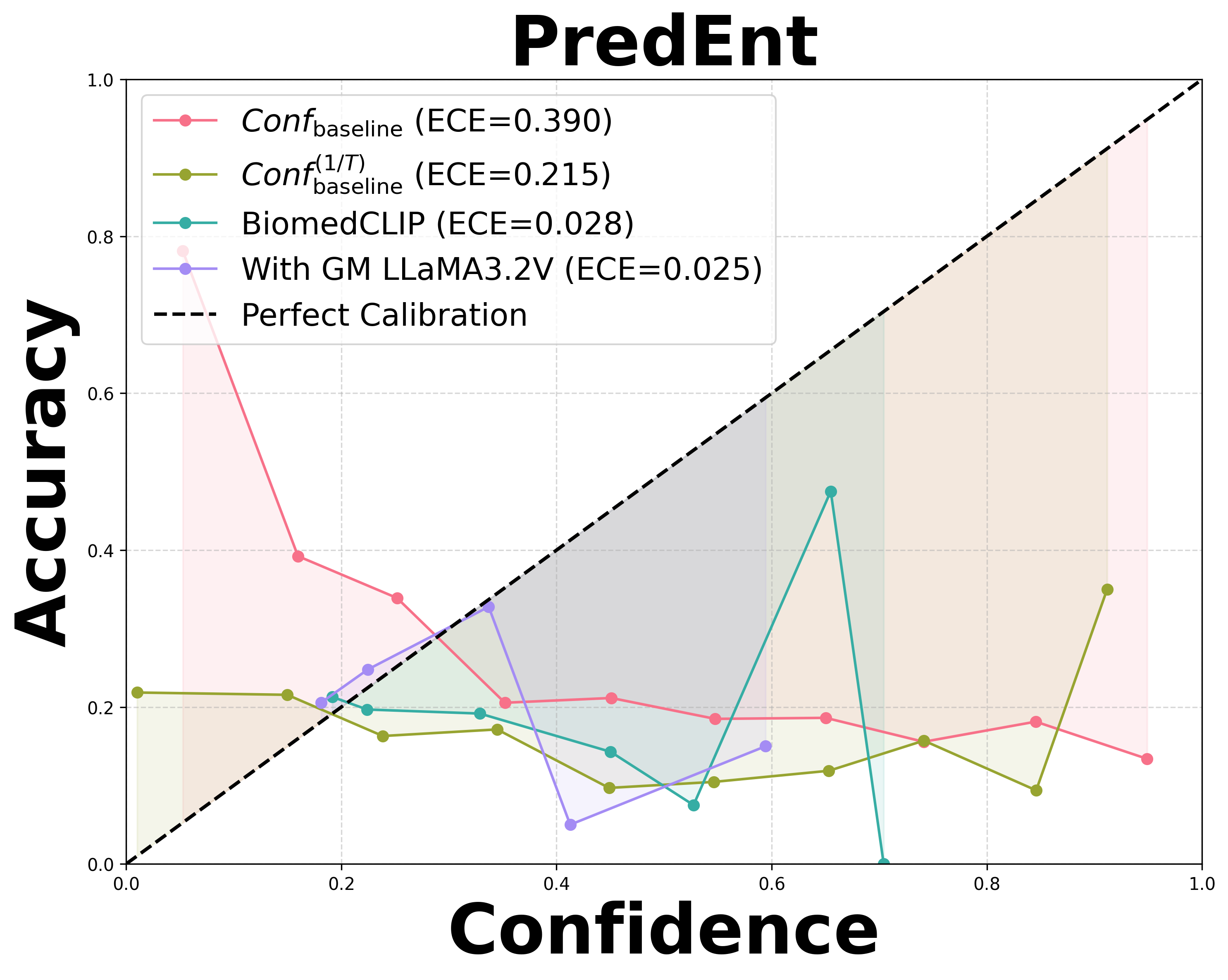}
        \label{fig:pred_entropy}
    \end{subfigure}
    \hfill
    \begin{subfigure}[b]{0.45\textwidth}
        \includegraphics[width=\textwidth]{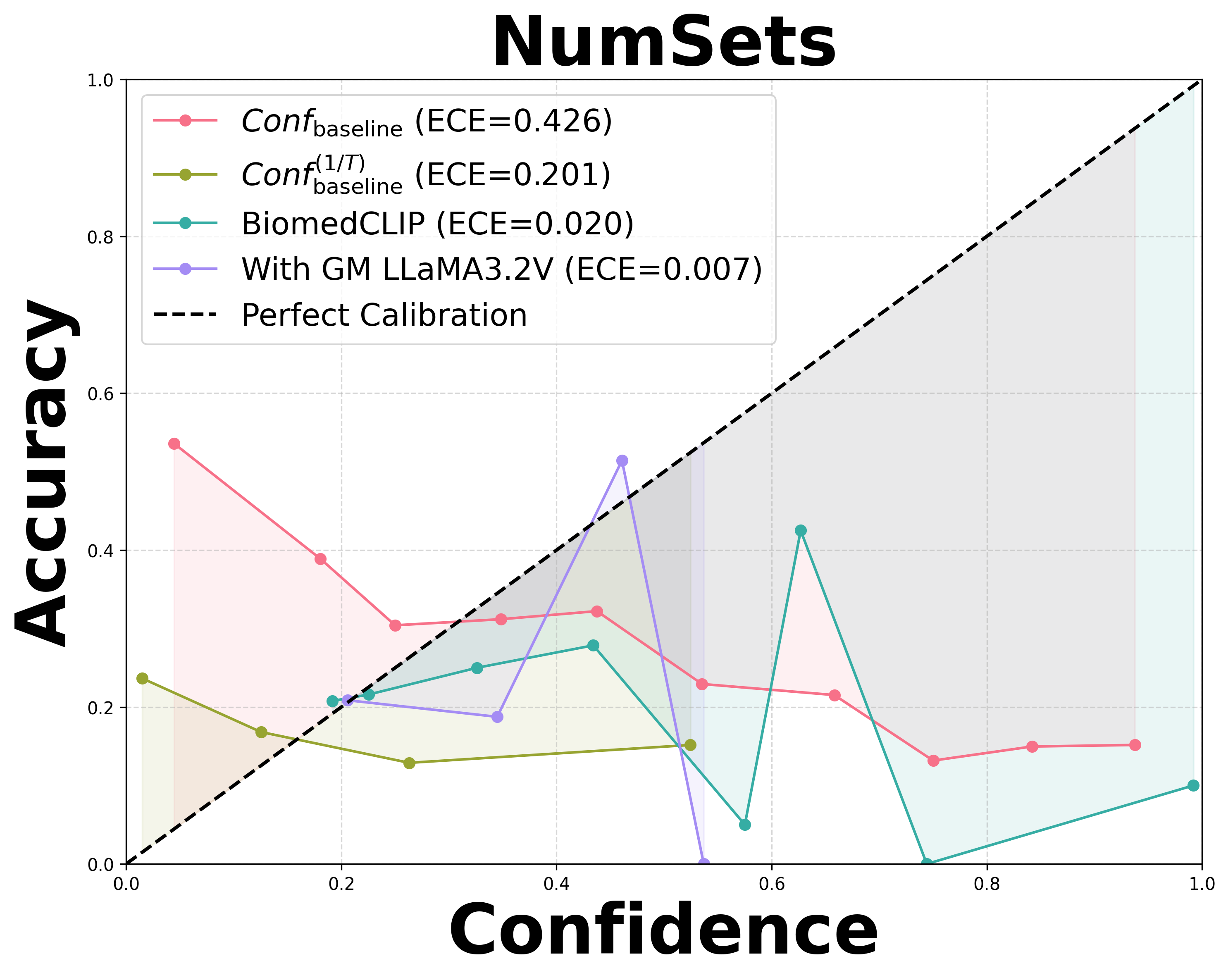}
        \label{fig:num_clusters}
    \end{subfigure}
    \caption{Reliability Diagrams for UQ by the two self-consistency baselines (PredEnt, NumSets) of LLaVA-Med on Slake. Each diagram shows plots and ECE for the confidence reported by the baseline $Conf_{baseline}$, its calibrated version $Conf_{baseline}^{(1/T)}$, and the proposed approach~\eqref{our_proposal} for calibration with different grounding models.}
    \label{fig:reliability_diagrams_slake}
\end{figure}

\subsection{ECE Results across different runs \& Hyperparameter Configurations}

\begin{table}[htbp]
\centering
\begin{tabular}{llcccc}
\toprule
\textbf{Baseline} & \textbf{Grounding Model} & \textbf{Mean ECE} & \textbf{Variance ECE} & \textbf{Mean T} & \textbf{Mean C} \\
\midrule
LexSim & Baseline     & 0.1692 & 0.00001 & NA   & NA  \\
LexSim & Temp. Scaled Baseline & 0.0813 & 0.00001 & 1.7  & NA  \\
LexSim & GroundedSAM          & 0.0524 & 0.00001 & 0.7  & 0.5 \\
LexSim & Qwen-VL      & 0.0606 & 0.00001 & 1.1  & 0.5 \\
LexSim & LLaMA 3.2V   & 0.0449 & 0.00001 & 0.3  & 0.5 \\
LexSim & CLIP         & 0.0453 & 0.00001 & 0.7  & 0.5 \\
\midrule
NumSets & Baseline     & 0.4096 & 0.00001 & NA   & NA  \\
NumSets & Temp. Scaled Baseline & 0.2465 & 0.00001 & 5.1  & NA  \\
NumSets & GroundedSAM          & 0.1432 & 0.0001 & 0.3  & 0.5 \\
NumSets & Qwen-VL      & 0.1249 & 0.0001 & 0.5  & 0.5 \\
NumSets & LLaMA 3.2V   & 0.1265 & 0.00001 & 0.3  & 0.5 \\
NumSets & CLIP         & 0.1434 & 0.00001 & 0.5  & 0.5 \\
\midrule
SemEnt & Baseline      & 0.1080 & 0.00001 & NA   & NA  \\
SemEnt & Temp. Scaled Baseline  & 0.1078 & 0.00001 & 0.9  & NA  \\
SemEnt & GroundedSAM           & 0.0355 & 0.00001 & 2.9  & 0.2 \\
SemEnt & Qwen-VL       & 0.0728 & 0.00001 & 0.9  & 0.5 \\
SemEnt & LLaMA 3.2V    & 0.0290 & 0.0001 & 0.3  & 0.4 \\
SemEnt & CLIP          & 0.0383 & 0.0001 & 4.7  & 0.1 \\
\midrule
PredEnt & Baseline     & 0.4445 & 0.0001 & NA   & NA  \\
PredEnt & Temp. Scaled Baseline & 0.1904 & 0.0002 & 9.7  & NA  \\
PredEnt & GroundedSAM          & 0.1145 & 0.0001 & 0.7  & 0.5 \\
PredEnt & Qwen-VL      & 0.1187 & 0.0001 & 2.7  & 0.5 \\
PredEnt & LLaMA 3.2V   & 0.0964 & 0.0001 & 0.3  & 0.5 \\
PredEnt & CLIP         & 0.1165 & 0.0001 & 0.9  & 0.5 \\
\bottomrule
\end{tabular}
\caption{The table reports the mean Expected Calibration Error (ECE), variance of ECE across five random splits over test/validation sets, across different grounding models and baselines for the Slake dataset. and the corresponding average values of the temperature scaling parameter (T) and confidence threshold (C). Results are presented for four baselines—LexSim, NumSets, SemEnt, and PredEnt—each evaluated with a vanilla baseline, temperature-scaled baseline, and multiple vision-language grounding models including GroundedSAM, Qwen-VL, LLaMA 3.2V, and CLIP.}
\label{tab:expanded_ece_comparison}
\end{table}

\newpage
\begin{table}[htbp]
\centering
\begin{tabular}{llcccc}
\toprule
\textbf{Baseline} & \textbf{Grounding Model} & \textbf{Mean ECE} & \textbf{Variance ECE} & \textbf{Mean T} & \textbf{Mean C} \\
\midrule
LexSim & Baseline         & 0.0308 & 0.00001 &   NA    &  NA     \\
LexSim & Temp. Scaled Baseline    & 0.0312 & 0.00001 & 0.9   &    NA  \\
LexSim & BiomedCLIP       & 0.0040 & 0.00001 & 0.1   & 0.2  \\
LexSim & LLaMA 3.2V       & 0.0134 & 0.00001 & 0.5   & 0.1  \\
\midrule
NumSets & Baseline        & 0.4263 & 0.00001 &    NA   &   NA   \\
NumSets & Temp. Scaled Baseline    & 0.2015 & 0.00001 & 0.1   &  NA    \\
NumSets & BiomedCLIP      & 0.0070 & 0.00001 & 0.1   & 0.2  \\
NumSets & LLaMA 3.2V      & 0.0212 & 0.00001 & 0.1   & 0.2  \\
\midrule
SemEnt & Baseline         & 0.3764 & 0.00001 & NA      &   NA   \\
SemEnt & Temp. Scaled Baseline     & 0.2221 & 0.00001 & 0.1   &  NA    \\
SemEnt & BiomedCLIP       & 0.0080 & 0.00001 & 0.1   & 0.2  \\
SemEnt & LLaMA 3.2V       & 0.0098 & 0.00001 & 0.3   & 0.2  \\
\midrule
PredEnt & Baseline        & 0.3902 & 0.00001 &  NA     &   NA   \\
PredEnt & Temp. Scaled Baseline    & 0.2149 & 0.00001 & 0.1   &   NA   \\
PredEnt & BiomedCLIP      & 0.0069 & 0.00001 & 0.1   & 0.2  \\
PredEnt & LLaMA 3.2V      & 0.0138 & 0.00001 & 0.1   & 0.2  \\
\bottomrule
\end{tabular}
\caption{The table reports the mean Expected Calibration Error (ECE), variance of ECE across five random splits over test/validation sets, and the average values of temperature (T) and confidence threshold (C) used during calibration. Each baseline—LexSim, NumSets, SemEnt, and PredEnt—is evaluated under three settings: a vanilla baseline, a temperature-scaled baseline, and two grounding models: BiomedCLIP (a biomedical domain-specific model) and LLaMA 3.2V}
\label{tab:ece_comparison}
\end{table}

\end{document}